%% file: main.tex
\pdfoutput=1

\documentclass[11pt]{article}

\usepackage[]{EMNLP2023}

\usepackage{times}
\usepackage{latexsym}
\usepackage[T1]{fontenc}

\usepackage{svg}
\usepackage[utf8]{inputenc}
\usepackage{xspace}
\usepackage{microtype}
\usepackage{inconsolata}
\usepackage{colortbl}
\usepackage{amsmath}
\usepackage{amssymb}
\usepackage{array}
\usepackage{pifont}
\usepackage{tabularx}
\usepackage{adjustbox}
\usepackage{enumitem}
\usepackage{multirow}
\usepackage{dsfont}
\usepackage{booktabs}
\usepackage{MnSymbol}
\usepackage{scalerel}
\usepackage{mathrsfs}
\usepackage{graphicx}
\usepackage{changes}
\usepackage{xcolor,soul}
\usepackage{makecell}
\usepackage{ulem}
\usepackage{graphicx,calc}
\usepackage{bbm}
\usepackage{lipsum}
\usepackage[caption=false]{subfig}
\usepackage{algorithm}
\usepackage{algpseudocode}

\usepackage[most]{tcolorbox}

\newtcbox{\hlprimarytab}{on line, rounded corners, box align=base, colback=white!10,colframe=white,size=fbox,arc=3pt, before upper=\strut, top=-2pt, bottom=-4pt, left=-2pt, right=-2pt, boxrule=0pt}
\newtcbox{\hlprimarytabg}{on line, rounded corners, box align=base, colback=gray!10,colframe=white,size=fbox,arc=3pt, before upper=\strut, top=-2pt, bottom=-4pt, left=-2pt, right=-2pt, boxrule=0pt}
\newtcbox{\hlsecondarytab}{on line, box align=base, colback=red!10,colframe=white,size=fbox,arc=3pt, before upper=\strut, top=-2pt, bottom=-4pt, left=-2pt, right=-2pt, boxrule=0pt}

\definecolor{darkgreen}{RGB}{0,100,0}
\definecolor{darkred}{RGB}{200,0,0}
\definecolor{lightgreen}{RGB}{228,253,227}
\definecolor{lightred}{RGB}{252,231,234}
\definecolor{lightyellow}{RGB}{250,253,191}
\definecolor{lightblue}{RGB}{230,240,254}
\definecolor{white}{RGB}{255,255,255}

\newcommand{\whethermath}[1]{\ifmmode{#1}\else{$#1$}\fi}

\newcommand{\phz}{\ifmmode\phantom{0}\else$\phantom{0}$\fi}

\usepackage{cleveref}

\title{Understanding GUI Agent Localization Biases through Logit Sharpness}

\author{Xingjian Tao\textsuperscript{1}, Yiwei Wang\textsuperscript{3}, Yujun Cai\textsuperscript{4}, 
Zhicheng Yang\textsuperscript{1}, Jing Tang\textsuperscript{1,2}, \\
\textsuperscript{1}The Hong Kong University of Science and Technology (Guangzhou),\\ ~\textsuperscript{2}The Hong Kong University of Science and Technology,
~\textsuperscript{3}University of California, Merced,
\\~\textsuperscript{4}The University of Queensland
\\
\texttt{taoxj2001@outlook.com, wangyw.evan@gmail.com, jingtang@ust.hk}\\
}

\begin{document}
\maketitle
\begin{abstract}
Multimodal large language models (MLLMs) have enabled GUI agents to interact with operating systems by grounding language into spatial actions. Despite their promising performance, these models frequently exhibit hallucinations—systematic localization errors that compromise reliability. We propose a fine-grained evaluation framework that categorizes model predictions into four distinct types, revealing nuanced failure modes beyond traditional accuracy metrics. To better quantify model uncertainty, we introduce the Peak Sharpness Score (PSS), a metric that evaluates the alignment between semantic continuity and logits distribution in coordinate prediction. Building on this insight, we further propose Context-Aware Cropping, a training-free technique that improves model performance by adaptively refining input context. Extensive experiments demonstrate that our framework and methods provide actionable insights and enhance the interpretability and robustness of GUI agent behavior.

\end{abstract}

\input{sec/1introduction}
\input{sec/2related}
\input{sec/3hallucination}

\input{sec/4logits}

\input{sec/5method}

\input{sec/conclusion}



\bibliography{main}
\bibliographystyle{acl_natbib}

\appendix
\input{sec/appendix}

\end{document}

%% file: sec/1introduction.tex
\section{Introduction}
Recent progress in large language models (LLMs; \citealt{touvron2023llama1,vicuna2023,falcon40b,MosaicML2023Introducing,touvron2023llama,openaichatgptblog,bardclaudeblog}) has greatly advanced natural language understanding and generation. However, their reliance on purely textual inputs and outputs limits their applicability in perceptual and interactive tasks.
To address this, multi-modal large language models (MLLMs) have emerged by incorporating visual inputs alongside text. Models such as Flamingo~\cite{alayrac2022flamingo}, Gemini~\cite{team2023gemini}, and Qwen-VL~\cite{bai2023qwen,wang2024qwen2,bai2025qwen2} enable more comprehensive reasoning across modalities, supporting tasks like visual question answering, image captioning, and document understanding where both language and vision are crucial.

Building on recent advances in LLMs and MLLMs, researchers have turned to GUI agents—intelligent systems capable of autonomously operating graphical user interfaces via perception, reasoning, and action. These agents must interpret screen layouts, understand task instructions (often in natural language), and generate accurate sequences of interface actions such as clicks or keystrokes. Effective GUI agents require the integration of language understanding, visual perception, and action planning in dynamic environments.

A key challenge lies in accurately identifying interaction targets on the interface. General-purpose MLLMs often fall short in GUI-specific tasks, particularly in predicting precise operation coordinates. To overcome this, recent approaches adopt training-based pipelines that enhance the agent's capabilities through continued pre-training on large auxiliary datasets, followed by the integration or adaptation of neural modules tailored for GUI tasks. This foundation enables more effective fine-tuning on smaller, domain-specific datasets, improving precision and robustness in real-world GUI interactions.

Existing GUI agent models are typically pre-trained or fine-tuned on large-scale datasets of interface operations. However, they often underperform on tasks that are trivial for human users. Moreover, these models exhibit weak localization accuracy for rarely seen or uncommon icons, which hinders their ability to generalize across diverse interface environments.

In this study, we systematically analyze the forms of hallucination exhibited by existing GUI agent models. These hallucinations often manifest as inaccurate or implausible predictions during icon localization, especially in cases involving unfamiliar interface elements. To facilitate a controlled investigation, we introduce a novel icon library containing symbols that are semantically clear and visually distinctive, yet rarely encountered in common human-computer interaction scenarios. These icons are designed to integrate naturally into GUI layouts while providing new challenges for model generalization.

To better characterize and quantify hallucinations in GUI localization tasks, we propose formal definitions and a taxonomy of hallucination types, along with corresponding classification algorithms. This framework allows for a more precise evaluation of model behavior when interacting with both familiar and unfamiliar interface elements.

In parallel, we examine the distribution of logit scores produced by GUI agent models during coordinate prediction. Unlike conventional natural language tasks such as question answering, GUI localization tasks typically involve output tokens that represent numeric values (e.g., x and y coordinates). These tokens exist in an ordered space, where semantic proximity is inherently meaningful. For instance, when predicting the token ``6'', surrounding tokens like ``5'' and ``7'' are expected to have higher logits due to their closeness in both numerical and spatial terms, whereas tokens such as ``1'' or ``9'' are more distant in this context. This characteristic provides a unique opportunity to study structured output spaces and the nature of model uncertainty in GUI interaction tasks.

In addition, informed by our analysis, we introduce an image cropping strategy that effectively reduces hallucinations and enhances the coordinate prediction performance of existing GUI agent models.

This work makes several key contributions:
\begin{itemize}[topsep=4pt, itemsep=0pt]
    \item We present a systematic analysis of hallucination behaviors in GUI agents and introduce a formal taxonomy for localization errors;
    \item We analyze logit distributions in coordinate predictions, uncovering structured uncertainty unique to GUI-based tasks;
    \item We propose a targeted image cropping strategy that enhances coordinate prediction performance in existing GUI agent models.
\end{itemize}

%% file: sec/2related.tex
\section{Related Work}
\paragraph{Multi-modal language models} Multi-modal language models (MLLMs) integrate visual and textual information, enabling joint reasoning across modalities~\cite{bai2023qwen, wang2024qwen2,bai2025qwen2, alayrac2022flamingo, team2023gemini, ma2023llm, yang2023dawn, liu2023visual, li2023llava}. Early models like VilBERT~\cite{lu2019vilbert} and VisualBERT~\cite{li2019visualbert} extended BERT to handle vision-language tasks. More recent architectures such as Flamingo~\cite{alayrac2022flamingo}, OFA~\cite{wang2022ofa}, and BLIP-2~\cite{li2023blip} leverage pretrained vision encoders and language models to achieve strong performance on image captioning, VQA, and document understanding. Models like Kosmos-2~\cite{peng2023kosmos} and Qwen-VL~\cite{bai2023qwen,wang2024qwen2,bai2025qwen2} further enhance grounding and layout understanding, which are particularly relevant for structured GUI environments. However, these general-purpose MLLMs still face challenges in precise spatial reasoning and action prediction required by GUI tasks.

\paragraph{GUI agents} GUI agents~\cite{nguyen2024gui,zhang2024large, cheng2024seeclick,lin2024showui, lu2024omniparser}, are designed to interact with graphical user interfaces through visual perception, natural language understanding, and action planning. Early generalist agents such as Gato~\cite{reed2022generalist} demonstrated multitask capabilities across robotic control, games, and web interfaces, but lacked fine-grained spatial grounding. 
Recent approaches have expanded GUI agent capabilities through stronger visual grounding and multimodal reasoning. WebGUM~\cite{furuta2023multimodal} introduces a hierarchical planning framework that combines LLMs with execution modules and perceptual grounding. OmniParser~\cite{lu2024omniparser} uses auxiliary visual models to mark the position of elements in the operation interface, improving the performance of GPT-4V in GUI agent tasks. ShowUI~\cite{lin2024showui} and UGround~\cite{gou2025uground} synthesize a large amount of training data for training efficient GUI agent models.
Despite progress, many agents still struggle with localization of unfamiliar UI elements and suffer from hallucination-like errors, especially in low-resource or distribution-shifted settings. This motivates continued research into robust visual-language grounding, more diverse pretraining data, and better uncertainty modeling in GUI environments.

%% file: sec/3hallucination.tex
\section{Investigating Hallucinations in GUI Agent Models}

\begin{figure}[!tb]
	\centering
	\includegraphics[width=1\linewidth]{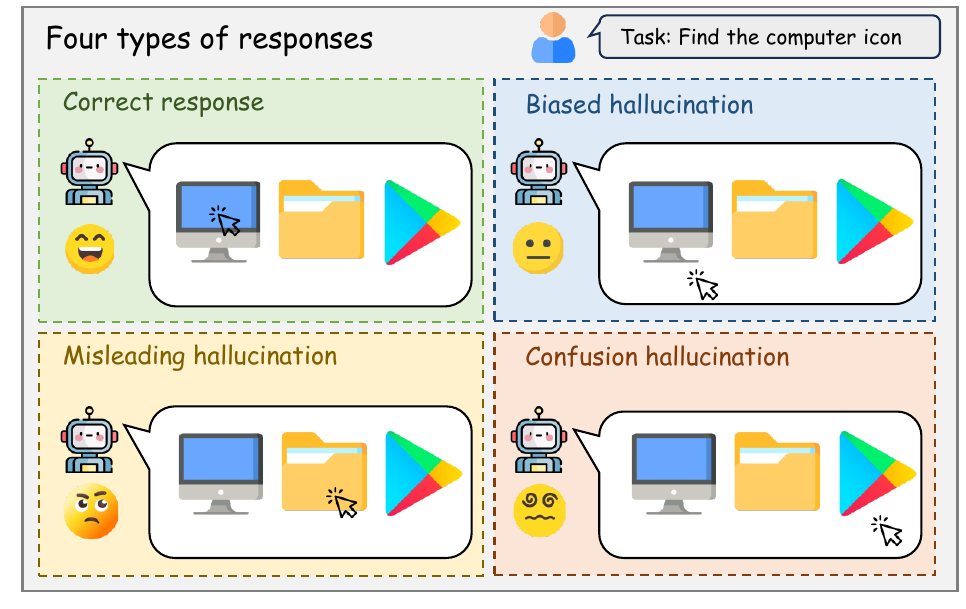}
	\caption{
Illustration of the four response types in GUI icon localization tasks. 
\label{fig:case}}
\end{figure}

Most existing benchmarks emphasize task success and interaction accuracy with GUI elements, but often overlook a nuanced analysis of hallucinations—systematic or implausible errors arising during model interaction. These evaluations typically rely on screenshots and provide only the ground truth bounding box, neglecting the potential influence of surrounding interface elements.

To address this limitation, we design a series of controlled experiments aimed at systematically analyzing hallucination behaviors in GUI agents. Furthermore, we introduce a dedicated classification algorithm to categorize hallucinated outputs, enabling deeper insight into error types and their underlying causes in GUI-based localization and interaction tasks.

Specifically, we categorize model responses into four types: correct responses, biased hallucinations, misleading hallucinations, and confusion hallucinations.

\subsection{How far are the model’s predictions from the ground truth?}
As presented in \Cref{tab:dis}, our empirical analysis demonstrates that, even in the presence of hallucinated predictions, the output coordinates generated by GUI agent models frequently lie in close proximity to the ground-truth region. 

This is especially evident in biased hallucinations, where the model correctly identifies the target icon semantically but fails in precise localization. The spatial clustering of such errors near the target indicates structured uncertainty rather than random noise.
These results expose the limitations of binary accuracy metrics, which overlook the nuanced spatial behaviors of GUI agents. This underscores the need for refined evaluation methods that quantify spatial proximity and provide interpretable diagnostics of model failure.

\begin{table}[t!]
\centering
\begin{tabular}{@{}>{\centering\arraybackslash}m{4.5cm} >{\centering\arraybackslash}m{2.5cm}@{}}
\toprule
\textbf{Evaluation Condition} & \textbf{Proportion (\%)} \\
\midrule
Correct response & 75.9 \\
Relative distance $<$ 0.05 & 84.5 \\
Relative distance $<$ 0.10 & 87.4 \\
Relative distance $<$ 0.20 & 90.9 \\
Relative distance $<$ 0.30 & 93.9 \\
\bottomrule
\end{tabular}
\vspace{0.5em}
\caption{Proportion of ShowUI-2B~\cite{lin2024showui} model predictions falling within various distance thresholds from the ground-truth bounding box. Evaluated on the ScreenSpot~\cite{cheng2024seeclick} dataset.}
\label{tab:dis}
\end{table}

\begin{figure*}[!tb]
	\centering
	\includegraphics[width=1\linewidth]{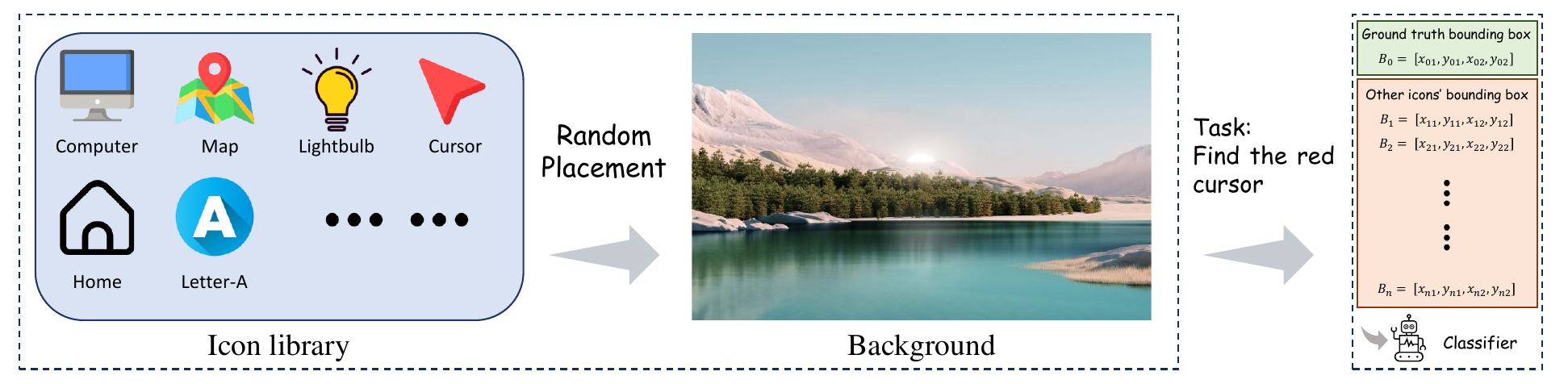}
	\caption{
Illustration of the experimental procedure for classifying responses generated by GUI agent models. A Windows desktop wallpaper is used as the background, onto which a set of icons is randomly placed. The bounding box of each icon is recorded and subsequently used to categorize the model's predicted coordinates according to the classification algorithm described in~\Cref{alg:class}.
\label{fig:icon}}
\end{figure*}
\subsection{Experiments Setup}
\paragraph{Baseline models and Benchmarks} 
We evaluate two novel and efficient GUI agent models, ShowUI-2B~\cite{lin2024showui} and UGround-V1-2B~\cite{gou2025uground}, both trained on the Qwen2-VL~\cite{wang2024qwen2} framework. 
All icons used in our experiments are obtained from publicly available open-source icon libraries, and the background images are default wallpapers from the Windows operating system. These assets are used solely for academic and non-commercial research purposes.

\subsection{Experiments design}
In prior benchmarks, the evaluation of GUI agents' click accuracy typically relies on determining whether the predicted coordinates fall within a predefined ground-truth bounding box. However, this binary protocol presents two significant limitations. First, it neglects the spatial distance between the predicted and actual positions, thus failing to capture the extent to which incorrect predictions deviate from the intended target. Second, it offers little insight into the underlying causes of hallucination errors—for example, whether the model was misled by visually similar or spatially adjacent elements.

Moreover, existing benchmarks often only provide the bounding box of the ground-truth target, without detailed annotations of other interface elements. This lack of contextual information makes it difficult to systematically investigate the origins and categories of hallucinations exhibited by GUI agent models. These limitations underscore the need for a more granular and interpretable evaluation framework tailored to GUI localization tasks.

To address these challenges, we design a controlled evaluation setting that incorporates a curated icon library composed of visually distinctive and semantically unambiguous icons with well-defined boundaries, as shown in \Cref{fig:icon}. In each experiment, a subset of icons is randomly placed on a synthetically generated GUI background, with one icon designated as the target. Since the exact positions and bounding boxes of all icons are known, we are able to conduct a fine-grained analysis of model predictions, focusing on spatial deviations and confusion behaviors.



\subsection{Taxonomy of GUI localization hallucinations}

Although prior research has primarily focused on task completion accuracy, our study reveals that the coordinates predicted by GUI agent models in operating system tasks can be further categorized into several distinct subtypes of responses. This finer-grained classification enables a deeper understanding of model behavior beyond simple success or failure.

Given a task instruction and a corresponding GUI image, a GUI agent model generates a coordinate pair \([x, y]\). The ground-truth target is defined by a bounding box 
$B_0 = [x_1, y_1, x_2, y_2]$.
A prediction is considered correct if it satisfies the condition $x_1 < x < x_2 \quad\text{and}\quad y_1 < y < y_2$ meaning the point lies within the ground-truth region.

We denote the set of all icons \(I\) placed on the GUI background as a set of bounding boxes $B = \{B_1, B_2, \dots, B_n\}$
each corresponding to a distinct icon with known coordinates. This setup allows us to determine whether a model's prediction corresponds to a wrong but plausible icon (e.g., visually similar or nearby), or is entirely spurious.

The specific response classification procedure is formalized in ~\Cref{alg:class}, which outlines how predicted coordinates are categorized based on their spatial relationship to \(B_0\) and other icons in the set \(B\). Specifically, we classify the responses of GUI agent models into the following categories:

\begin{itemize}
    \item \textbf{Correct response}: The predicted coordinates fall within the ground-truth bounding box \(B_0\).
    \item \textbf{Biased hallucination}: The prediction is close to the ground-truth region but lies outside of \(B_0\), suggesting a minor spatial deviation.
    \item \textbf{Misleading hallucination}: The coordinates fall near another icon’s bounding box \(B_i \in B\), indicating the model was misled by a visually or semantically similar distractor.
    \item \textbf{Confusion hallucination}: The output does not correspond to any identifiable icon, and the prediction appears unrelated to any meaningful visual element.
\end{itemize}

\begin{algorithm}[t!]
\caption{Classify GUI Agent Response Based on Bounding Box Distance \label{alg:class}}
\begin{algorithmic}[1]
\Require Predicted point $(x, y)$, ground-truth box $B_0 = [x_1, y_1, x_2, y_2]$, icon boxes $B = \{B_1, B_2, \dots, B_n\}$, distance threshold $\tau$
\Ensure Response category: \texttt{Correct}, \texttt{Biased}, \texttt{Misleading}, or \texttt{Confusion}

\If{$x_1 < x < x_2$ \textbf{and} $y_1 < y < y_2$}
    \State \Return \texttt{Correct response}
\EndIf

\State Compute distance $d \gets$ \Call{PointToBoxDistance}{$x, y, B_0$}
\If{$d < \tau$}
    \State \Return \texttt{Biased hallucination}
\EndIf

\For{each $B_i \in B$}
    \State Compute $d_i \gets$ \Call{PointToBoxDistance}{$x, y, B_i$}
    \If{$d_i < \tau$}
        \State \Return \texttt{Misleading hallucination}
    \EndIf
\EndFor

\State \Return \texttt{Confusion hallucination}

\vspace{1em}
\Function{PointToBoxDistance}{$x, y, [x_1, y_1, x_2, y_2]$}
    \State $dx \gets \max(x_1 - x, 0, x - x_2)$
    \State $dy \gets \max(y_1 - y, 0, y - y_2)$
    \State \Return $\sqrt{dx^2 + dy^2}$
\EndFunction
\end{algorithmic}
\end{algorithm}

\begin{figure}[!tb]
	\centering
	\includegraphics[width=1\linewidth]{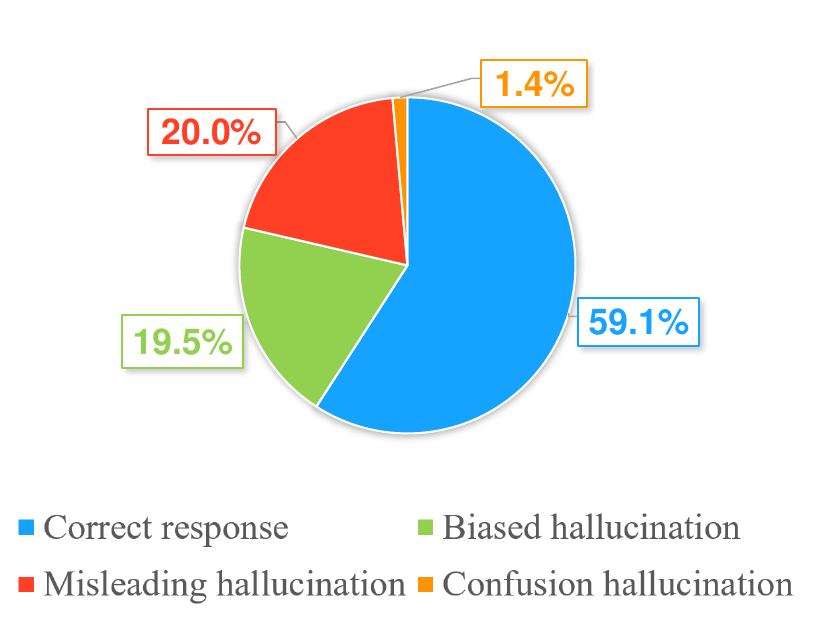}
	\caption{
Distribution of response types for different GUI agent models on the icon finding task. Percentages indicate the proportion of predictions belonging to each category as defined by our response classification framework. 
\label{fig:class}}
\end{figure}

As shown in~\Cref{fig:class}, our experimental results indicate that GUI agent models exhibit lower performance when tasked with locating icons that rarely appear in the operating interface, despite the relative simplicity of these tasks and the minimal requirement for semantic understanding. Notably, a substantial proportion of the observed errors fall into the category of biased hallucinations, where the model correctly identifies the target icon at a semantic or perceptual level but produces coordinate predictions that are slightly offset from the ground-truth region. In contrast, misleading hallucinations occur when the model either misinterprets the intended meaning of the icon or is misled by visually or spatially similar distractors, resulting in more pronounced localization errors.

Moreover, our analysis shows that confusion hallucinations constitute only a small fraction of the errors in the icon localization task. This suggests that GUI agent models are generally capable of extracting and leveraging meaningful visual elements from the interface, even when precise localization is imperfect.

In summary, different types of hallucinations exhibit distinct behavioral patterns, highlighting the limitations of binary classification schemes that simply judge coordinate predictions as either correct or incorrect. Such coarse-grained evaluation fails to capture the nuanced characteristics of model responses. To advance the development of more robust GUI agent models, there is a pressing need for finer-grained indicators capable of differentiating between hallucination types. In particular, we seek metrics that can effectively distinguish among various error modes, thereby enabling targeted analysis and method-specific improvements. We elaborate on this direction in the following section.


%% file: sec/4logits.tex
\section{New Metric: Peak Sharpness Score (PSS)}

\begin{figure*}[!tb]
	\centering
	\includegraphics[width=1\linewidth]{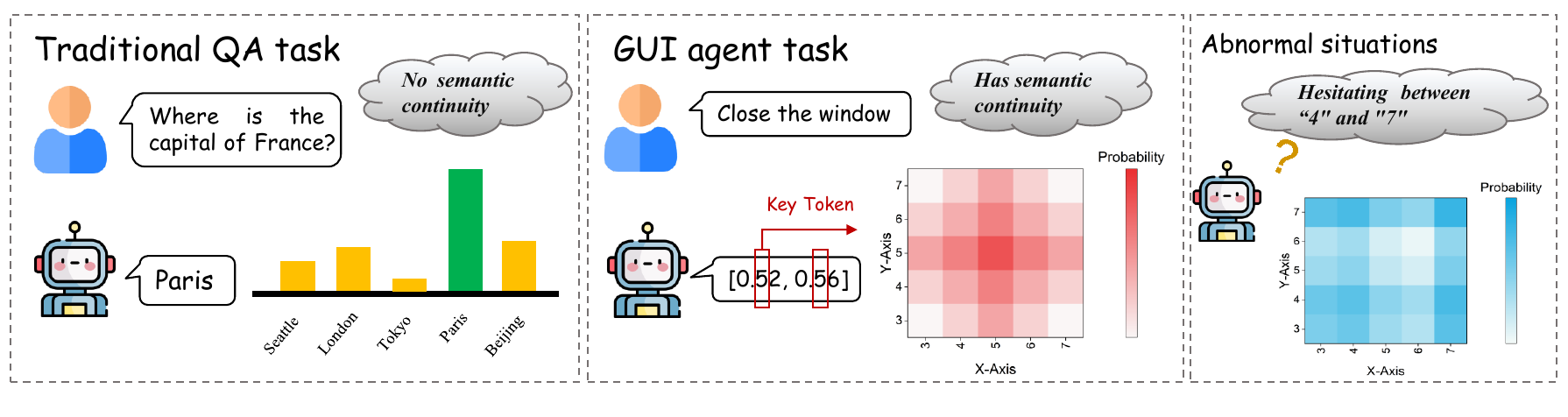}
	\caption{
Comparison of logits score distributions between GUI agent tasks and traditional question answering tasks. In GUI agent tasks, the model outputs coordinate values, where numeric tokens exhibit semantic continuity. In anomalous cases, a mismatch between this semantic continuity and the continuity of the logits distribution often indicates increased model uncertainty or confusion.
\label{fig:metric}}
\end{figure*}

While the previous section analyzed hallucination types and spatial errors in GUI agent predictions, the models’ confidence in these outputs remains insufficiently studied. Traditional metrics like accuracy and perplexity do not capture confidence effectively, especially in multimodal tasks involving spatial reasoning. To address this gap, we introduce a novel confidence-oriented metric tailored for GUI agents, offering finer-grained insight into model certainty and supporting more precise failure analysis.

\subsection{Analysis of Logit Distribution in GUI Agent Tasks}
Unlike traditional natural language tasks such as knowledge-based question answering, GUI localization tasks require models to output tokens representing numerical values—specifically, 
x and y coordinates. These tokens lie within an ordered, continuous space where semantic proximity directly reflects spatial closeness. For instance, when the model predicts the token ``6'', it is expected that nearby tokens like ``5'' and ``7'' will also receive relatively high logit scores, while distant tokens such as ``1'' or ``9'' should be less probable. This structured output space offers a unique opportunity to assess model uncertainty in a more interpretable and task-relevant way.

However, our experimental observations show that this expected pattern is frequently violated, particularly in cases identified as Misleading Hallucinations and Confusing Hallucinations. In such scenarios, the logit distribution does not exhibit the anticipated continuity, suggesting a breakdown in the model’s spatial grounding or confidence calibration.
To address this, we propose a new metric—Peak Sharpness Score (PSS)—which quantifies the alignment between semantic continuity and the shape of the logits distribution. 

\subsection{Definition of new metric}
\paragraph{Definition of key token}  The GUI agent model produces coordinate outputs as strings, such as ``[0.71, 0.23]'', with coordinates normalized to the range [0, 1]. Within these strings, certain tokens, such as ``7'' and ``2'' in the example, predominantly determine the coordinate values. We define these tokens, which critically influence the numerical representation of coordinates, as key tokens. The subsequent analysis will focus on these key tokens.

\paragraph{Definition of Semantic Continuity.}
We define \textit{semantic continuity} as the property of a sequence of tokens whose semantic representations vary smoothly and predictably in the embedding space. Let $T = \{t_1, t_2, \dots, t_n\}$ be a sequence of tokens, and $f: T \rightarrow \mathbb{R}^d$ be an embedding function mapping each token to a $d$-dimensional semantic vector $\mathbf{v}_i = f(t_i)$. Semantic continuity holds if the similarity between adjacent embeddings remains high, i.e.,
\[
\text{cos}(f(t_i), f(t_{i+1})) \approx 1 \quad \text{for all } i,
\]
and the embedding differences are approximately constant:
\[
f(t_{i+1}) - f(t_i) \approx f(t_i) - f(t_{i-1}).
\]
This implies a near-linear progression in embedding space. For example, numerical tokens such as ``1'', ``2'', and ``3'' typically exhibit semantic continuity. In contrast, tokens representing entities such as ``Paris'', ``London'', and ``Beijing'' lack such linearity due to their discrete and context-dependent meanings.

\paragraph{Motivation} In GUI agent tasks, token sequences representing coordinate values exhibit inherent semantic continuity, where neighboring tokens correspond to spatial proximity. Our experiments reveal that correct predictions typically preserve a smooth and continuous logits distribution aligned with this semantic structure—a property we define as the consistency between semantic continuity and logits distribution. To quantify this behavior and better assess model confidence, we introduce a novel metric that incorporates two key factors: (1) the peak logit value associated with the selected token during greedy decoding, and (2) the smoothness of logits across semantically adjacent tokens.

\paragraph{Definition of Peak Sharpness Score}
We propose the \textbf{Peak Sharpness Score (PSS)} as a novel metric to quantify the structural sharpness and symmetry of the logits distribution over coordinate tokens. This score is designed to reflect the model's confidence in coordinate-based predictions, particularly in GUI agent tasks where output tokens represent numerical values from a discrete and ordered space, such as $\{0, 1, \dots, 9\}$.

Let $V = [v_0, v_1, \dots, v_9]$ denote the logits vector over the 10 discrete coordinate tokens. We first identify the index $p = \arg\max(V)$ of the peak logit value $m = v_p$, which represents the model's most probable prediction.

If the peak lies at the boundary (i.e., $p = 0$ or $p = 9$), we cannot evaluate the symmetry of the distribution, as one side is missing. In such cases, we compute the average slope on the available side:
\[
s = \frac{1}{9} \sum_{i=0}^{8} (v_{i+1} - v_i)
\]
The final score is computed as:
\[
\text{PSS} = 2 \cdot |s| \cdot m
\]
This formulation amplifies edge cases while ensuring compatibility with the full formula used for interior peaks.

If the peak lies strictly inside the sequence ($0 < p < 9$), we split the logits into two segments: a \textit{left segment} $L = \{v_0, \dots, v_p\}$ and a \textit{right segment} $R = \{v_p, \dots, v_9\}$. For each segment, we compute the average slope:
\[
a_{\text{left}} = \frac{1}{p} \sum_{i=0}^{p-1} (v_{i+1} - v_i), \quad 
a_{\text{right}} = \frac{1}{9 - p} \sum_{i=p}^{8} (v_{i+1} - v_i)
\]
We then compute a weighted average of the absolute slopes, where the weights correspond to the segment lengths:
\[
w = \frac{p \cdot |a_{\text{left}}| + (9 - p) \cdot |a_{\text{right}}|}{9}
\]
Finally, the Peak Sharpness Score is defined as:
\[
\text{PSS} = C \cdot w \cdot m
\]
where $C$ is a normalization constant (empirically set to $C=4.5$), and $m$ is the peak logit value. This multiplication ensures that the metric reflects not only the sharpness of the distribution but also the confidence the model assigns to its top prediction.

A higher PSS indicates a more concentrated and confident (unimodal) distribution, while lower scores correspond to flatter or ambiguous predictions, which are often associated with hallucination behaviors.

\begin{table*}[ht]
\centering
\small
\begin{tabular}{lcccc}
\toprule
\textbf{Model} & \textbf{Correct} & \textbf{Biased Hallucination} & \textbf{Other Response}  \\
\midrule

Qwen2-VL-7B & 0.34 ± 0.28 & 0.34 ± 0.29 & 0.32 ± 0.28\\
ShowUI-2B & 0.59 ± 0.33 & 0.54 ± 0.33 & 0.40 ± 0.31  \\
UGround-V1-2B & 0.52 ± 0.30 & 0.43 ± 0.30 & 0.25 ± 0.24  \\
\bottomrule
\end{tabular}
\vspace{0.5em}
\caption{Peak Sharpness Score (PSS) of different GUI agent models across response categories. Values are reported as mean ± standard deviation.}
\label{tab:pss_comparison}
\end{table*}

\begin{table*}[ht]
\centering
\begin{tabular}{lccc}
\toprule
\textbf{Group Comparison} & \textbf{Biased vs. Correct} & \textbf{Other vs. Correct} & \textbf{Biased vs. Other} \\
\midrule
\textbf{Significance (p < 0.05)} & $\times$ & $\checkmark$ & $\checkmark$ \\
\bottomrule
\end{tabular}
\vspace{0.5em}
\caption{Pairwise significance test results on Peak Sharpness Score (PSS) across different response types. A check mark ($\checkmark$) indicates a statistically significant difference, while a cross ($\times$) indicates no significant difference.}
\label{tab:significance}
\end{table*}

\subsection{Experiment}
We evaluate two novel and efficient GUI agent models, ShowUI-2B~\cite{lin2024showui} and UGround-V1-2B~\cite{gou2025uground}, both trained on the Qwen2-VL~\cite{wang2024qwen2} framework. 
For testing, we utilize ScreenSpot~\cite{cheng2024seeclick}, a GUI agent evaluation dataset encompassing diverse operating interface types, including desktop, mobile, and web environments.

Experimental results reveal significant differences in PSS across various types of model responses. These differences indicate that PSS effectively captures variations in the confidence and structure of the logits distribution, providing a useful signal for distinguishing between correct predictions and different forms of hallucinations.

Our experimental results demonstrate that the Peak Sharpness Score (PSS) for correct predictions is significantly higher than that for incorrect responses. On the ScreenSpot benchmark, only the bounding box of the ground-truth target is provided, while the bounding boxes of other interface elements are not available. As a result, it is not feasible to distinguish between misleading and confusion hallucinations on this dataset; these two error types are therefore grouped together as other incorrect responses.

Notably, among the incorrect samples, biased hallucinations exhibit an average PSS that is closer to that of correct responses than to other error types. Furthermore, t-test analysis reveals that the difference in PSS between correct and biased hallucination samples is not statistically significant. This suggests that although biased hallucinations are technically incorrect, the model's confidence and output structure in these cases remain comparable to that of correct predictions.

\subsection{Analysis}

Our experimental findings highlight two key insights:

\begin{enumerate}
    \item \textbf{Biased hallucinations produce logits distributions resembling those of correct responses}, suggesting the model correctly identifies the target but suffers from minor coordinate deviations. This issue is especially common with small icons, where precise localization is more difficult.

    \item \textbf{Misleading and confusion hallucinations yield significantly lower Peak Sharpness Scores (PSS)}, indicating the model fails to recognize the correct element. These errors often arise from semantic or visual similarity between interface icons, leading to incorrect target selection rather than localization inaccuracy.

\end{enumerate}

%% file: sec/5method.tex
\begin{table*}[htbp]
\centering
\small
\begin{tabular}{lccccccc}
\toprule
\textbf{Model} & Desktop icon & Desktop text & Mobile icon & Mobile text & Web icon & Web text & \textbf{Avg.} \\
\cmidrule(lr){1-8}
\cmidrule(lr){1-8}
Qwen2-VL-2B & 10.7 & 14.9 & 14.8 & 22.7 & 11.6 & 26.5 & 17.7 \\
Qwen2-VL-7B & 28.5 & 42.7 & 48.5 & 74.4 & 30.1 & 46.9 & 47.7 \\
Qwen2.5-VL-3B & 5.7 & 18.0 & 2.6 & 6.2 & 0.5 & 0.9 & 5.4 \\
Qwen2.5-VL-7B & 15.0 & 47.4 & 10.0 & 42.1 & 4.4 & 12.6 & 22.7 \\
ShowUI-2B & 61.1 & 76.3 & 75.5 & 92.3 & 63.6 & 81.7 & 75.1 \\
UGround-V1-2B & 65.7 & 88.7 & 72.0 & 89.4 & 68.9 & 81.3 & 77.7 \\
\cmidrule(lr){1-8}

Qwen2-VL-7B + Crop & 37.1 & 45.9 & 58.5 & 80.6 & 36.4 & 50.0 & 53.9 \\
ShowUI-2B + Crop & 63.6 & 88.6 & 72.9 & 93.0 & 65.0 & 82.1 & 79.0 \\
UGround-V1-2B + Crop & 67.1 & 89.2 & 66.4 & 91.6 & 69.9 & 83.9 & 79.1 \\
\bottomrule
\end{tabular}
\caption{Comparison of localization accuracy before and after applying the proposed cropping strategy. The cropping method centers the predicted point and retains $\alpha=0.8$ of the original image size. The accuracy improvements across different models demonstrate the effectiveness of spatially focused input preprocessing.}
\label{tab:gui_model_comparison}
\end{table*}

\begin{table*}[htbp]
\centering
\small

\begin{tabular}{lccc}
\toprule
\textbf{Model} & \textbf{Correct} & \textbf{Biased Hallucination} & \textbf{Other Response} \\
\cmidrule(lr){1-4}
\cmidrule(lr){1-4}
ShowUI-2B & $0.59 \pm 0.33$ & $0.54 \pm 0.33$ & $0.40 \pm 0.31$ \\
ShowUI-2B + Crop & $0.64 \pm 0.33$ & $0.61 \pm 0.32$ & $0.47 \pm 0.33$ \\
UGround-V1-2B & $0.52 \pm 0.30$ & $0.43 \pm 0.30$ & $0.25 \pm 0.24$ \\
UGround-V1-2B + Crop & $0.66 \pm 0.31$ & $0.46 \pm 0.28$ & $0.46 \pm 0.33$ \\
\bottomrule

\end{tabular}
\caption{Comparison of Peak Sharpness Scores (PSS) before and after cropping for different response types across two GUI agent models.}
\label{tab:comparsion_score}
\end{table*}

\section{Context-Aware Cropping}
\subsection{Methodology}
Our analysis reveals that biased hallucinations often stem from slight coordinate deviations, especially when the target icon occupies a small portion of the screen. To mitigate this, we propose cropping the screen into smaller regions centered around the query. This reduces spatial complexity and encourages the model to focus on local visual cues, improving alignment between semantic understanding and coordinate prediction.

To reduce spatial ambiguity and mitigate biased hallucinations in GUI agent predictions, we propose Context-Aware Cropping that refocuses the model’s attention on a local region surrounding the target. Given a full-screen image $I$ of size $H \times W$ and a predicted coordinate $(x, y) \in [0, 1]^2$, we define a cropped region $I'$ with dimensions $\alpha H \times \alpha W$, where $0 < \alpha < 1$, and aim to position the target coordinate as close as possible to the geometric center of the cropped region.

The cropping region is computed as follows:
\begin{align*}
    x_{\text{start}} &= \min\left(\max\left(xW - \frac{\alpha W}{2},\, 0\right),\, W - \alpha W\right), \\
    y_{\text{start}} &= \min\left(\max\left(yH - \frac{\alpha H}{2},\, 0\right),\, H - \alpha H\right).
\end{align*}

This ensures that the crop stays within image boundaries while centering the predicted point as much as possible. In our experiments, we set the crop ratio $\alpha = 0.8$ to retain sufficient contextual information while reducing peripheral distractions.

\subsection{Experiment and analysis}
To evaluate the effectiveness of our proposed cropping strategy, we conducted a comparative experiment on multiple GUI agent models, measuring their performance before and after applying the cropping operation. 
Experimental results show that this simple yet effective preprocessing step consistently improves localization accuracy across various models. For instance, the accuracy of \texttt{ShowUI-2B} improved from 75.1\% to 79.0\% after applying cropping. Notably, the improvement is especially pronounced for cases classified as biased hallucinations, where the original prediction was close to the correct region but slightly off-center. This confirms our hypothesis that cropping reduces irrelevant visual context and helps the model focus more accurately on the intended GUI elements.

As shown in~\Cref{tab:comparsion_score}, to quantitatively assess the impact of spatially focused inputs on model confidence, we evaluated changes in the Peak Sharpness Score (PSS) following the application of our cropping strategy. As previously defined, PSS captures the alignment between semantic continuity and the continuity of the model's logit distribution, thereby serving as a task-relevant indicator of predictive certainty. Upon cropping the input image to center around the predicted coordinate—while retaining an $\alpha$ proportion of the original dimensions—we observed a consistent improvement in PSS across all evaluated models. This indicates that the removal of peripheral visual clutter enhances the model’s ability to assign concentrated probability mass to semantically adjacent tokens. The results suggest that localizing visual context not only improves spatial accuracy but also strengthens the internal confidence structure of the model, offering a principled mechanism for reducing uncertainty in GUI-based localization tasks.

%% file: sec/conclusion.tex
\section{Conclusion}
We present a controlled experimental framework for evaluating GUI agent models via an icon localization task with fully annotated interface elements. This setup enables fine-grained classification of model responses, revealing diverse hallucination patterns beyond binary accuracy.

To better characterize model uncertainty, we introduce the \textit{Peak Sharpness Score} (PSS), a metric that quantifies the alignment between semantic continuity of coordinate tokens and the smoothness of logit distributions. Empirical results show that different hallucination types correspond to distinct PSS profiles, offering insights into their underlying causes.

Additionally, we propose \textit{Context-Aware Cropping}, a simple yet effective method that improves localization performance and model confidence without retraining. Our findings highlight the importance of structured evaluation and representation-aware metrics in advancing robust GUI agents.

%% file: sec/appendix.tex
\section{Details of Prompts}
The zero-shot prompt used in this paper for ShowUI-2B is shown below.
\begin{tcolorbox}
System prompt:

``According to the image I provide, identify the relative coordinates of the specified object, with values ranging from 0 to 1. The output format must be [x, y], and do not output anything else.''

User prompt:

[Task]

\end{tcolorbox}

The zero-shot prompt used in this paper for other models is shown below.
\begin{tcolorbox}
User prompt:

Your task is to help the user identify the precise coordinates (x, y) of a specific area/element/object on the screen based on a description.

Description: [Task]

Answer:

\end{tcolorbox}

\section{More Experimental Data}

\begin{table}[ht]
\centering
\begin{tabular}{@{}lc@{}}
\toprule
\textbf{Response Type} & \textbf{Perplexity (↓)} \\
\midrule
Correct Response         & 1.12 \\
Biased Hallucination     & 1.17 \\
Misleading Hallucination & 1.30 \\
Confusing Hallucination  & 1.33 \\
\bottomrule
\end{tabular}
\vspace{0.5em}
\caption{Perplexity scores for different types of model responses. Lower perplexity indicates higher model confidence.}
\label{tab:perplexity}
\end{table}

\section{Case study}

We show examples of background images from the icon finding task, as well as sample demonstrations of the four types of responses. The output coordinates of the model are marked with blue dots.

\begin{figure*}[!tb]
	\centering
	\includegraphics[width=1\linewidth]{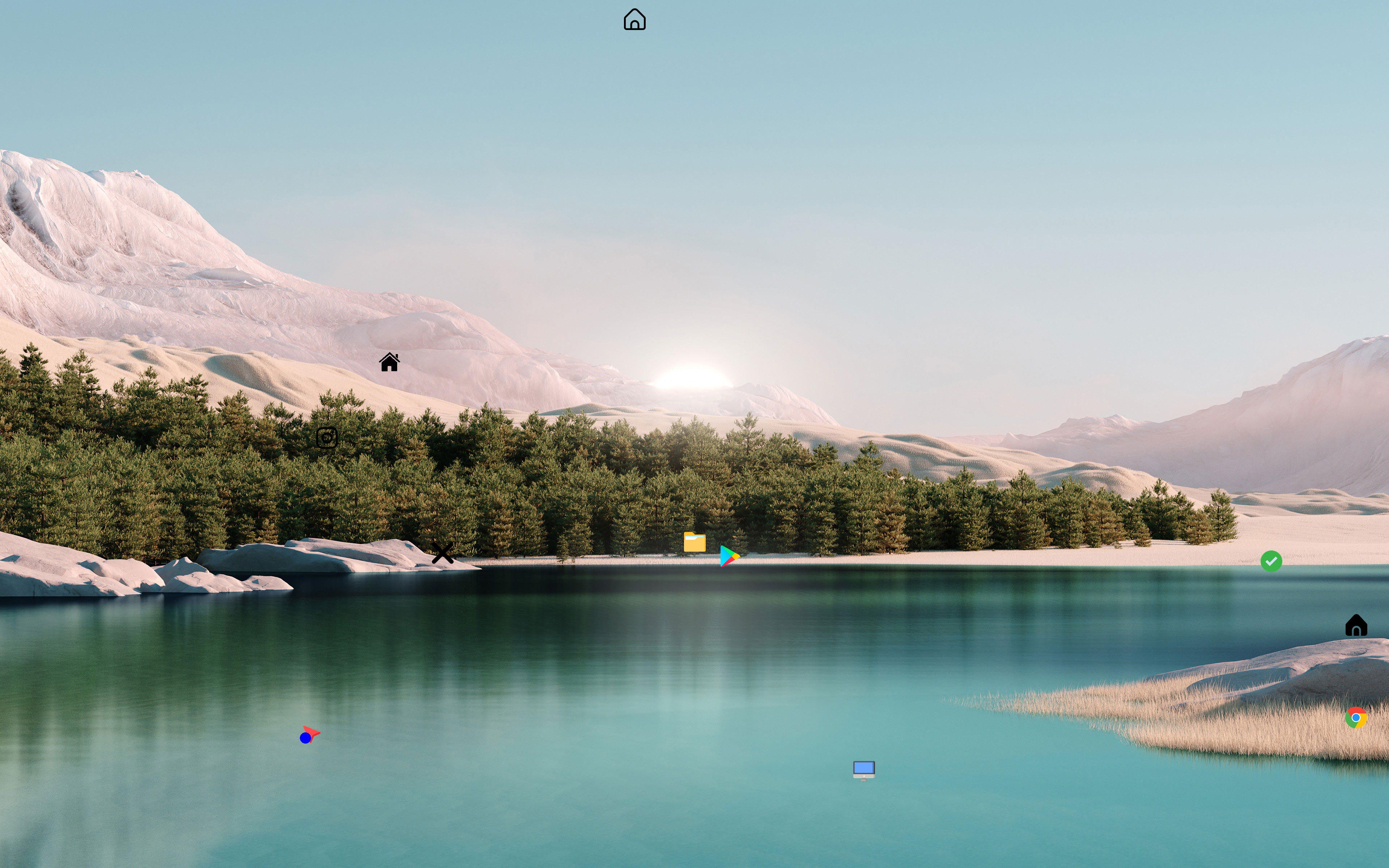}
	\caption{
Demonstration example of correct response.
\label{fig:icon1}}
\end{figure*}

\begin{figure*}[!tb]
	\centering
	\includegraphics[width=1\linewidth]{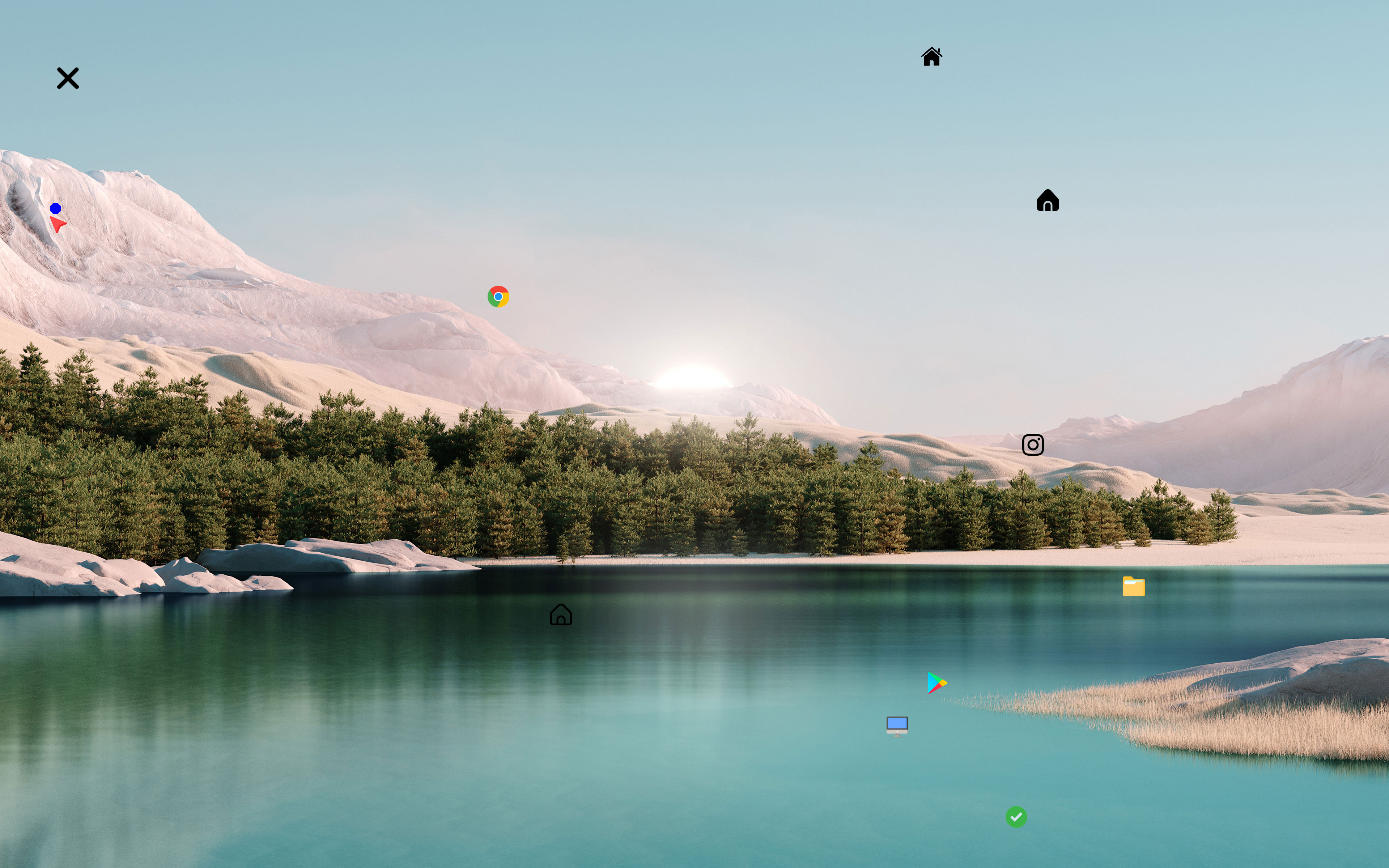}
	\caption{
Demonstration example of biased hallucination.
\label{fig:icon2}}
\end{figure*}

\begin{figure*}[!tb]
	\centering
	\includegraphics[width=1\linewidth]{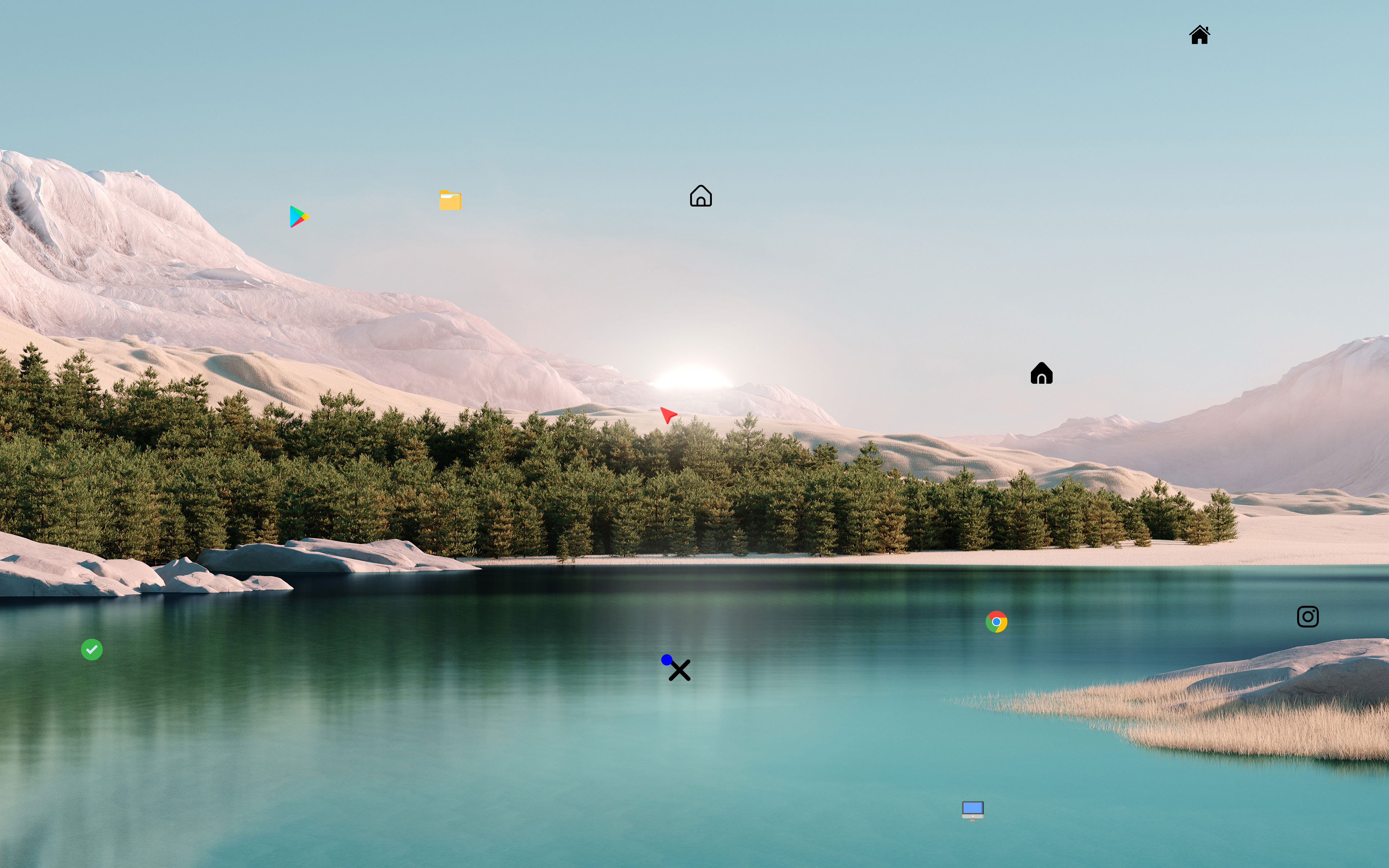}
	\caption{
Demonstration example of misleading hallucination.
\label{fig:icon3}}
\end{figure*}

\begin{figure*}[!tb]
	\centering
	\includegraphics[width=1\linewidth]{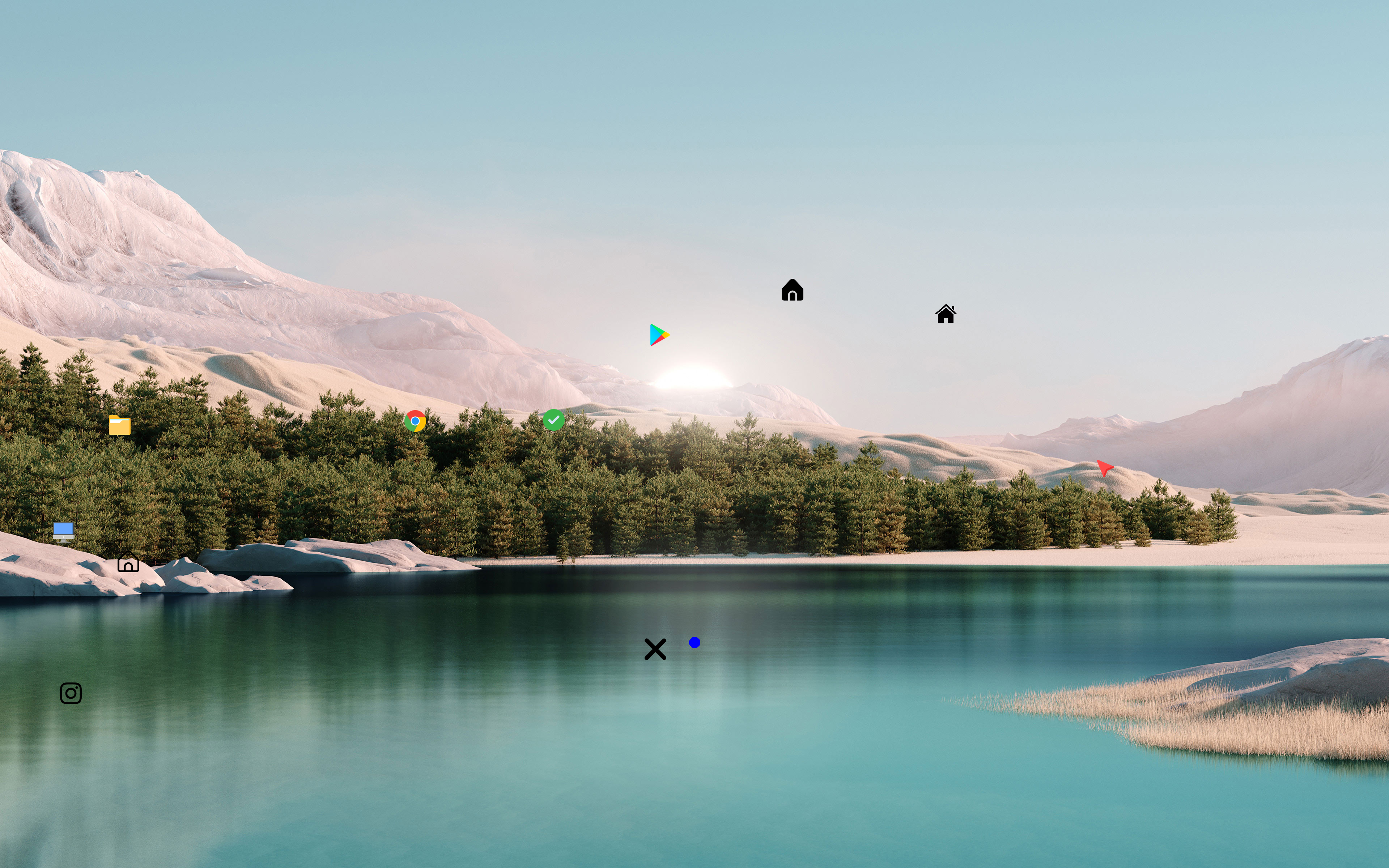}
	\caption{
Demonstration example of confusion hallucination.
\label{fig:icon4}}
\end{figure*}